\documentclass[conference,letterpaper]{IEEEtran}
\IEEEoverridecommandlockouts

\usepackage{etoolbox}
\makeatletter
\patchcmd{\@maketitle}{\centering}{\hrule\@height\z@\@depth\z@\vskip0.25in\centering}{}
  {\ClassWarningNoLine{IEEEtran}{Could not patch \string\@maketitle\space for the
   ICRA 1.0in first-page top margin -- check the title margin by hand}}
\makeatother

\usepackage{cite}
\usepackage{amsmath,amssymb,amsfonts}
\usepackage{algorithmic}
\usepackage{graphicx}
\usepackage{textcomp}
\usepackage{xcolor}
\usepackage{booktabs}
\usepackage{multirow}
\usepackage{url}
\usepackage[hidelinks]{hyperref}
\usepackage{stfloats}   % allow table*/figure* at page bottoms
\usepackage{placeins}   % \FloatBarrier

\def\BibTeX{{\rm B\kern-.05em{\sc i\kern-.025em b}\kern-.08em
    T\kern-.1667em\lower.7ex\hbox{E}\kern-.125emX}}

\newcommand{\sigt}{\ensuremath{\sigma_t}}
\newcommand{\sigf}{\ensuremath{\sigma_f}}
\newcommand{\Tint}{\ensuremath{\mathcal{T}}}

\newcommand{\hi}{{\scriptsize$\uparrow$}}          % higher is better
\newcommand{\lo}{{\scriptsize$\downarrow$}}        % lower is better
\newcommand{\tgt}[1]{{\scriptsize$\to\!#1$}}       % closer to #1 is better

\begin{document}

\title{CellSplat4D: PSF-Aware 4D Gaussian Splatting\\
for Sparse Robotic Live-Cell Imaging}

\author{
    \IEEEauthorblockN{Yingda Tao and Guoyu Lu}
    \IEEEauthorblockA{Indiana University Bloomington}
}

\maketitle

% ---------------------------------------------------------------------------
\begin{abstract}
A robotic microscope watching living cells cannot afford to look as often as it would like. Every volume it acquires costs photons the specimen does not get back, and time owed to other wells. What such a platform exists to produce is a record of individual cells through time: which cell is which from one volume to the next, and which cell divided into which two. Sampling sparsely breaks that record exactly where it matters, and the fault lies in the acquisition schedule rather than in the analysis software.

We fill the gaps by reconstructing them, fitting a 4D Gaussian model to whatever volumes the hardware could afford. The model is a cloud of light-emitting blobs, each carrying a position, a shape and a lifetime. Being continuous in time, it renders any missing volume on demand, decoupling how often the robot analyses from how often it can afford to look. The microscope's point-spread function is measured from the data rather than inherited from acquisition metadata or left to the optimizer, because metadata inflates it and the optimizer cannot recover it at all: a wider blur around a smaller blob fits the images equally well. Each blob's lifetime is stored in frames rather than as a fraction of the recording, so that it denotes a fixed duration on any sequence. Unmeasured timesteps are supervised at coarse scale by a 3D U-Net that predicts the intermediate volume directly and estimates no motion field, since a dividing cell becomes two and no motion describes that.

On two Cell Tracking Challenge sequences, a C.\ elegans embryo and a Chinese Hamster Ovarian (CHO), with fidelity scored per cell nucleus, our reconstruction holds the highest nucleus fidelity at every distance from an acquired frame, has the flattest decay across the gap, and best recovers focal planes it was never shown with graceful degradation across the gap.
\end{abstract}

% ---------------------------------------------------------------------------
\section{Introduction}

Robotic platforms now run live-cell experiments end to end, and the most
advanced decide what to image next from what they have just seen
\cite{conrad2011micropilot,mahecic2022event,pinkard2021pycromanager};
micro-robotic systems close a tighter loop still, acting on individual cells
under visual servoing \cite{sun2002biological,nelson2010microrobots}. All are
limited by the same resource. A volumetric acquisition costs stage travel and
photons the specimen does not get back, excess exposure changes the biology
rather than merely adding noise \cite{icha2017phototoxicity,laissue2017assessing},
and a platform serving many wells must divide that budget
\cite{huisken2004spim,chen2014lattice,royer2016adaptive}. So it samples
\emph{sparsely in time} and accepts the gaps.

\textbf{The gaps are where the robot's autonomy fails.} Linkers lose identities
across the jump, lineage graphs fragment at the divisions they exist to record,
and a controller choosing its next target reasons over a state it does not
have. These are documented failure modes as the sampling interval grows
\cite{ulman2017objective,maska2023ctc}, and the deficiency is in the
acquisition, not the tracker.

% ---------------------------------------------------------------------------
\begin{figure*}[!tb]
\centering
\includegraphics[width=\textwidth]{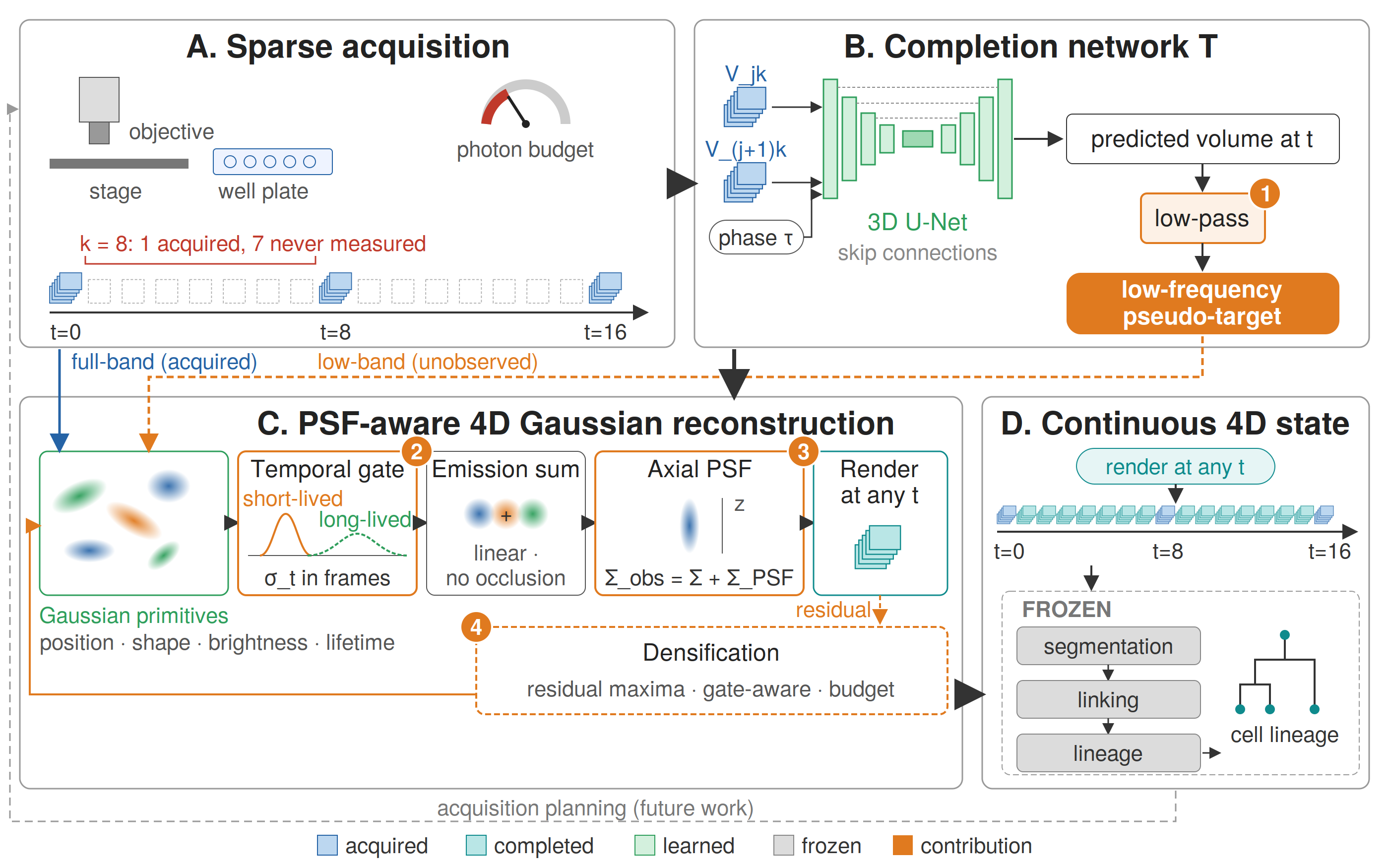}
\caption{We turn a sparsely sampled acquisition into a continuous-time
volumetric state. Because the state is queried by rendering, the rate at which
the robot's downstream modules run is decoupled from the rate at which its
hardware can afford to acquire.}
\label{fig:workflow}
\end{figure*}

Our response is to give the robot a \emph{continuous-time 4D state} of the
specimen, fitted to whatever the hardware could afford and queried by rendering
(Fig.~\ref{fig:workflow}). Three things follow. It \emph{decouples the analysis rate from the acquisition
rate}: every downstream module was designed for a densely sampled sequence, and
a rendered volume enters the pipeline where an acquired one would. It
\emph{produces a physical state rather than pixels}, primitives with positions,
extents and lifetimes that can be asked where an emitter is and how long it
persists, which interpolated images cannot supply. And it \emph{changes the
acquisition design space}: the sampling interval becomes a parameter the robot
trades rather than a constraint it lives under, which reframes what we ask of a
dynamic representation. Not whether it fits the acquired frames, since all do,
but how gracefully its per-nucleus fidelity \emph{decays} with distance from
the nearest acquired frame.
\textbf{Flatness, not peak accuracy, is what buys a longer acquisition
interval}, and it is the axis on which we evaluate.

Gaussian splatting \cite{kerbl20233dgs} and its dynamic extensions
\cite{yang2024deformable,duan20244drotor,li2024spacetime,
yang2024realtime4dgs,luiten2024dynamic} are the natural substrate: explicit,
fast to render, and exposing per-primitive temporal parameters that can be
inspected rather than probed. Designed for multi-view video of opaque surfaces
at full frame rate, three of their assumptions break under sparse volumetric
fluorescence. Their temporal gate is a \emph{fraction of the training window},
so a fixed initialization denotes a different duration on every sequence
(Sec.~\ref{sec:gate}). Their forward model treats axial blur as free or
defaulted, when it is a physical constant the default overstates and the
optimizer cannot recover (Sec.~\ref{sec:whypsf}, Sec.~\ref{sec:psf}). And
their densification statistics ignore the gate, diluting the growth signal in
proportion to window length (Sec.~\ref{sec:densify}). Each failure is silent:
training loss and full-image PSNR register none of them, and
Sec.~\ref{sec:psnrfails} shows PSNR ranks the fixes backwards.

We contribute \textbf{(1)} a completion framework that turns a sparse
acquisition into a continuous-time 4D state, scored on the decay profile of
nucleus fidelity across the gap rather than on aggregate image quality
(Sec.~\ref{sec:exp}); \textbf{(2)} three changes that make 4D splatting work at
this sampling rate: frame-unit temporal gates (Sec.~\ref{sec:gate}), an axial
PSF pinned to a measured physical constant, against which both the metadata
default and a learned width fail (Sec.~\ref{sec:psf}), and gate-aware
residual-driven densification under an exact budget (Sec.~\ref{sec:densify});
and \textbf{(3)} an evaluation protocol of masked, nucleus-level metrics
(Sec.~\ref{sec:eval}).

% ---------------------------------------------------------------------------
\section{Related Work}

\subsection{Preliminaries: 3D and 4D Gaussian splatting}
\label{sec:prelim}

3D Gaussian splatting \cite{kerbl20233dgs} represents a scene as anisotropic
primitives, primitive $i$ a Gaussian $G_i$ of mean $\boldsymbol{\mu}_i$ and
covariance $\boldsymbol{\Sigma}_i$, projected under an EWA approximation of the
pinhole map \cite{zwicker2001ewa} and accumulated front-to-back over
depth-sorted primitives,
\begin{equation}
C(\mathbf{u}) \;=\; \sum_{i} \mathbf{c}_i\,\alpha_i(\mathbf{u})
\prod_{j<i}\big(1-\alpha_j(\mathbf{u})\big),
\label{eq:alpha}
\end{equation}
with $\alpha_i = o_i G'_i$ for opacity $o_i$ and $G'_i$ the projected 2D
Gaussian. Its appeal here is that every term is an explicit, physically
interpretable parameter, not a network activation.

Dynamic extensions add time in one of two ways. \emph{Deformation-field}
methods \cite{pumarola2021dnerf,yang2024deformable,wu20244dgaussians} warp a
canonical primitive set with an MLP, often over a low-rank spacetime
factorization \cite{cao2023hexplane,fridovichkeil2023kplanes}. \emph{Native-4D}
methods
\cite{yang2024realtime4dgs,duan20244drotor} instead lift each primitive into
spacetime with a temporal centre $\mu_{t,i}$ and extent $\sigt{}_{,i}$, so that
marginalizing at time $t$ scales it by a \emph{temporal gate}
\begin{equation}
w_i(t) \;=\; \exp\!\Big(\!-\frac{(t-\mu_{t,i})^{2}}{2\,\sigt{}_{,i}^{2}}\Big).
\label{eq:gate}
\end{equation}
We build on the native-4D family because $\sigt{}_{,i}$ states how long the
model believes each emitter persists, which is what a sparse-acquisition
analysis needs to inspect and where Sec.~\ref{sec:gate} finds a failure.

\subsection{Why the forward model must carry the PSF}
\label{sec:whypsf}

Equation~\eqref{eq:alpha} encodes two assumptions that fail for a fluorescence
microscope. First, fluorophores are self-luminous and, in thin specimens,
neither occlude one another nor vary with viewing direction, so the
occlusion-aware composite collapses to a linear accumulation of emission,
\begin{equation}
I(\mathbf{x}) \;=\; \sum_i e_i\,G_i(\mathbf{x}),
\label{eq:emission}
\end{equation}
with $e_i$ a scalar emission weight. This removes depth sorting and view
dependence, making the rendered quantity a \emph{measurement} of integrated
intensity rather than a colour, the same reduction X-ray splatting makes
\cite{cai2024xgaussian}.

Second, the instrument is not a pinhole: the detector records the emission
field convolved with the point-spread function $h$,
\begin{equation}
\hat{I} \;=\; I * h,
\qquad
h \approx \mathcal{N}\!\big(\mathbf{0},\,
\mathrm{diag}(\sigma_{xy}^2,\sigma_{xy}^2,\sigf^{2})\big),
\label{eq:psf}
\end{equation}
where the axial width $\sigf$ far exceeds the lateral one and is \emph{a
property of the objective, not of the specimen}. Since convolving two Gaussians
sums their covariances, an emitter of true extent $\boldsymbol{\Sigma}_i$ is
observed with extent
\begin{equation}
\boldsymbol{\Sigma}_i^{\mathrm{obs}}
\;=\; \boldsymbol{\Sigma}_i + \boldsymbol{\Sigma}_{\mathrm{PSF}}.
\label{eq:sum}
\end{equation}

This has two consequences. A model omitting $h$ absorbs
$\boldsymbol{\Sigma}_{\mathrm{PSF}}$ into $\boldsymbol{\Sigma}_i$, recovering
the specimen convolved with the optics and inflated anisotropically in $z$. One
that includes $h$ but leaves $\sigf$ free inherits a non-identifiability
instead: shrinking every $\boldsymbol{\Sigma}_i$ while growing
$\boldsymbol{\Sigma}_{\mathrm{PSF}}$ leaves $\hat{I}$ unchanged on every
supervised plane, and only planes the model never saw break the tie. Pinning
$\sigf$ removes that degenerate direction and lets $\boldsymbol{\Sigma}_i$ be
read as a statement about the specimen.

GaussianPile \cite{gaussianpile} already encodes a finite-thickness PSF in its
projection operator, using the theoretical value. Our renderer inherits that
operator and lifts its primitives to 4D under Eq.~\eqref{eq:gate}; what we add
concerns the \emph{value} of $\sigf$ inside it, where Sec.~\ref{sec:psf} finds
failure modes on both sides of the theoretical choice.

\subsection{Dynamic scene representations}
Neural radiance fields \cite{mildenhall2020nerf,muller2022instantngp}
established volumetric inverse rendering before splatting displaced them on
speed. We compare against a deformation-field representative
\cite{yang2024deformable} and a radiative 4D variant built for sparse-view
angiography \cite{liu20244drgs}, which shares the emission-only regime
of~\eqref{eq:emission}. Feature-based \cite{li2024spacetime} and persistent
\cite{luiten2024dynamic} spacetime splatting both assume dense capture, as does
every method here. Sparsity has been attacked in the \emph{angular} domain by
few-shot splatting \cite{zhu2024fsgs}; ours is \emph{temporal}, the missing
quantity being the specimen's state between measurements rather than an
unobserved viewing angle. Our contribution is accordingly not a new dynamic
representation but an account of which of their components are load-bearing
when most frames are absent.

\subsection{Temporal completion}
Video frame interpolation is dominated by flow-based architectures that
estimate a motion field and warp along it
\cite{jiang2018superslomo,huang2022rife,reda2022film}. That design rests on
brightness constancy, which a dividing nucleus violates: it brightens and
becomes two objects where there was one, so no displacement field explains the
pair. Our completion network $\Tint$ is therefore a \emph{3D
U-Net} \cite{ronneberger2015unet,cicek20163dunet} regressing the intermediate
volume directly, baselined against the alternative it declines to be
\cite{farneback2003twoframe,teed2020raft}.

\subsection{Microscopy reconstruction and evaluation}
Content-aware restoration \cite{weigert2018care} established that learned
priors transfer across fluorescence modalities but can hallucinate structure
that image-level metrics reward. Putting the PSF inside the reconstruction
inverts the classical deconvolution arrangement, where it instead drives an
iterative restoration of the acquired volume \cite{sage2017deconvolutionlab2};
the constant supplied is the same, and so is the requirement that it be right.
Ours is the standard Gaussian approximation to the widefield axial response
\cite{zhang2007gaussianpsf,bornwolf1999principles}. Sequences and foreground
masks follow the Cell Tracking Challenge protocol
\cite{ulman2017objective,maska2023ctc}. Image-level SSIM/PSNR
\cite{wang2004ssim} is reported only for completeness;
Sec.~\ref{sec:psnrfails} shows it ranks methods backwards here.

% ---------------------------------------------------------------------------
\section{Problem Setting}

\subsection{Acquisition model}

A specimen is imaged as a time series of $z$-stacks. Under stride $k$ the robot
acquires only frames $t \in \{0, k, 2k, \dots\}$; the rest are never measured and
serve only for evaluation. Metrics are reported against $d$, the distance from
a held-out frame to the nearest acquired one.

We use two strides, $k=8$ and $k=4$. The main comparison is at $k=8$, the
sparsest we test and the regime in which nucleus fidelity in the middle of the
gap separates the methods most, with a second sequence reported at $k=4$. No
setting is tuned per sequence: the one hyper-parameter that depends on the
stride is the gate initialization, which we place at $k$ frames at both strides
(Sec.~\ref{sec:gate}).
Sequences are the C.\ elegans developing embryo and CHO sets of the Cell Tracking
Challenge \cite{ulman2017objective}.

A second holdout runs orthogonally to the first. A subset of $z$-planes is
withheld from supervision in every run, so that axial extrapolation can be
scored apart from fit: the model is fitted on the remaining planes of every
acquired volume and evaluated on the withheld ones. Wherever this paper says
\emph{held-out frame} it means the temporal holdout and \emph{held-out plane}
the axial one. Tables~\ref{tab:gate}--\ref{tab:gating} report the first;
Table~\ref{tab:cost} and the last column of Table~\ref{tab:ladder} report the
second.

The forward model splats anisotropic 3D Gaussians and convolves along $z$ with
a Gaussian axial PSF of width $\sigf$ (Sec.~\ref{sec:whypsf},
\cite{gaussianpile}), each primitive gated in time by Eq.~\eqref{eq:gate}.

\subsection{Why nucleus-level metrics}

The object the robot's downstream stack consumes is the nucleus, and it
occupies only about $8\%$ of the voxels. We therefore score per-nucleus
quantities on a frozen cross-arm panel: \emph{flux} (recovered integrated
intensity, normalized so that $1.0$ is the ground-truth nucleus),
\emph{sharpness}, and \emph{centroid offset} in absolute pixels. Unless stated
otherwise, numbers are the three-seed median over clean, quiescent nuclei;
where a contrast is sensitive to detector initialization we also report a
\emph{primed} variant seeded from the nearest acquired frame.

% ---------------------------------------------------------------------------
\section{Method}

\subsection{Temporal gates in frames, initialized wide}
\label{sec:gate}

A primitive's temporal extent $\sigt$ should mean how long this emitter
persists, in frames. Native-4D formulations store it as a fraction of the
training window, so the same number denotes a fraction of a frame on a short
clip and tens of frames on a long one: the parameter describes the experiment's
length, not the specimen. The damage lands where a sparse protocol operates,
since on long windows the conventional initialization is an order of magnitude
wider than the optimum and the optimizer does not recover. We therefore express
$\sigt$, its initialization and its learning rate in frames
(Table~\ref{tab:gate}).

% --- contributions.md Sec. 1.1 (report_densify.md) ---------------------------
\begin{table}[!tb]
\caption{The temporal gate stored as a fraction of the training window
(\emph{upstream}) and in frames (\emph{ours}), on a short recording and a long
one. Read \emph{down} a column to compare the two parameterizations at a fixed
window length; the bottom row is their difference. \textbf{The gain scales with
window length}, the signature of a parameter defined proportionally to that
length.}
\label{tab:gate}
\centering
\footnotesize
\begin{tabular}{lcc}
\toprule
Held-out nucleus flux\hi & \multicolumn{2}{c}{Training window}\\
\cmidrule(lr){2-3}
$\sigt$ stored as & $5$ frames & $100$ frames\\
\midrule
window fraction (upstream) & 0.934 & 0.478\\
\textbf{frames (ours)}     & \textbf{0.970} & \textbf{0.853}\\
\midrule
Gain (ours $-$ upstream)   & $+0.036$ & $\mathbf{+0.375}$\\
\bottomrule
\end{tabular}
\end{table}

The table separates two things. The absolute level is higher at five frames
under \emph{either} parameterization, because a five-frame window is a nearly
static problem in which one primitive can span the whole sequence. The
\emph{gain} is what the reparameterization is responsible for, and it grows
with window length, capacity and iteration count held fixed. A by-product
confirms the mechanism: the learned gate widths become window-independent and
vary with the \emph{trajectory} a nucleus follows.

Where the gate starts matters too, because the optimization is one-directional
in practice: \textbf{gates shrink easily and grow only with difficulty.} A gate
that is too wide produces visible misalignment and an unambiguous narrowing
gradient, whereas one that is too narrow can only widen if the primitive
simultaneously learns to move correctly over the longer interval, and
per-primitive velocity is not recoverable from sparse supervision. The rule
is practical rather than tuned: initialize \emph{above} the
expected optimum and let the optimizer descend. Every initial width at or above
the acquisition interval converges to the same solution while smaller ones
stall there permanently, so we initialize at the acquisition interval itself,
$k$ frames, at both strides rather than tuning a constant per sequence.

\subsection{A physical axial PSF}
\label{sec:psf}

The axial width $\sigf$ has three possible origins and only one is sound.

\emph{Neither defaulted nor learned.} The conventional data-preparation path
assigns $\sigf$ the $z$-step whenever no explicit value is passed, so every
prepared sequence carries an axial PSF equal to the plane spacing; across our
sequences that overstates the true blur by $1.9$--$2.3\times$, and the
reconstruction absorbs the excess by inflating every primitive in $z$. Leaving
$\sigf$ free fails the other way: extent and PSF width enter the observation
additively (Sec.~\ref{sec:whypsf}), so on supervised planes they are not
separately identifiable, and $\sigf$ duly drifts to $2.74\times$ the measured
value, buying $+0.176$ nucleus flux for $1.18$~dB of held-out axial PSNR. A
wider kernel spreads each primitive over more planes, which an
integrated-intensity score rewards and only unsupervised planes punish.

\emph{Measured, and checked against the optics.} We recover $\sigf$ per
sequence from the foreground axial autocorrelation and check it against the
widefield prediction $\mathrm{FWHM}_z \approx 2\lambda n/\mathrm{NA}^2$
\cite{zhang2007gaussianpsf,bornwolf1999principles}; the two agree to within a
few percent. The agreement rather than a win is the finding, because it
establishes that $\sigf$ is a \emph{constant of the instrument} at all, which
the metadata default and the learned value are not, and that licenses fixing
it. We measure rather than derive because the prediction assumes $\lambda$, $n$
and $\mathrm{NA}$ are correctly logged, and a disagreement would itself be
signal that a derived constant absorbs silently.

One consequence settles an acquisition-budget question. At the corrected width
the axial PSF spans barely more than a single plane, where the default had it
spanning more than two, so adjacent $z$-planes carry nearly independent
information and sparsifying the $z$-scan is close to free.
\textbf{The budget should be spent on time.}

\subsection{Capacity and supervision inside the gaps}
\label{sec:densify}

Splatting methods add capacity where the reconstruction is failing, deciding
where from an accumulated per-primitive gradient statistic already known to be
fragile on natural images \cite{zhang2024pixelgs}. Here it averages over every
primitive, including the majority whose gate is closed at the sampled timestep
and which contribute nothing but denominator, so the longer the window the more
the growth signal is diluted: densification behaves normally on a short clip
and silently stops on a long one. Restricting the statistic to gate-active
primitives repairs it.

The growth threshold is the second problem: inherited from natural-image
settings, it acts here as a step function rather than a dial, either never
firing or running to the primitive cap. Since a representation retrained per
experiment needs a predictable footprint, we replace it with an explicit
budget, as MCMC-based splatting does for a different reason
\cite{kheradmand2024mcmc}, place new primitives at residual maxima, and keep
capacity off the background with a significance floor. This buys a knob rather
than fidelity: the gate-aware fix alone restores growth.

The $k-1$ unobserved timesteps carry no supervision. We supply a low-frequency
pseudo-target from $\Tint$ as a blurred-domain loss: render and pseudo-target
are low-pass filtered before comparison, so $\Tint$ constrains only the band in
which it is trustworthy while high frequencies stay determined by the acquired
frames. $\Tint$ takes the two bracketing volumes and the phase between them and
regresses the intermediate volume directly, estimating no motion field, because
a division is not a displacement.

% ---------------------------------------------------------------------------
\section{Evaluation Protocol}
\label{sec:eval}

\subsection{Full-image PSNR fails on this task}
\label{sec:psnrfails}

We report three independent instances, all on identical checkpoints, held-out
frames and masks:
(1) Optical-flow warping and the frame-unit parameterization receive
\emph{opposite} rankings under nucleus flux and under PSNR. (2) Under the old
parameterization, at a gate width of $27$ frames and a nucleus flux of
$0.478$, temporal PSNR is \emph{higher} than after the fix. (3) Sweeping the
$\sigt$ initialization from $2$ to $24$ moves temporal PSNR by $0.42$~dB while
the nucleus flux gap across the same sweep is $25\%$.
The cause: $92\%$ of voxels are background, and warping transports the real,
pixel-independent sensor noise that fluorescence denoisers are built to
separate from signal \cite{krull2019noise2void}, which a pixel-wise metric
instead credits.

The remedy is to mask. \emph{Masked} PSNR is the ordinary
$10\log_{10}\!\big(\mathrm{MAX}^2/\mathrm{MSE}\big)$ with the error averaged over
a foreground mask $\Omega$ instead of the whole volume. That change is decisive
here: whole-volume PSNR is dominated by how faithfully empty space is
reproduced, so a reconstruction can lose a nucleus outright without moving it.
The same holds for SSIM, and a learned perceptual metric \cite{zhang2018lpips}
is no remedy either, its features being trained on natural photographs. Every
fidelity number is therefore masked: the nucleus metrics by
construction, and the $z$-PSNR columns of Table~\ref{tab:cost} over $\Omega$.
Whole-volume PSNR/SSIM \cite{wang2004ssim} is a sanity check.
$\Omega$ is the reference segmentation dilated by a few pixels \emph{within}
each $z$-plane, never across them: dilating in $3$D would let a held-out
plane's mask be defined partly by its supervised neighbours, scoring the axial
gap on planes the model was fitted to.

% ---------------------------------------------------------------------------
\begin{figure*}[!tb]
\centering
\includegraphics[width=\textwidth]{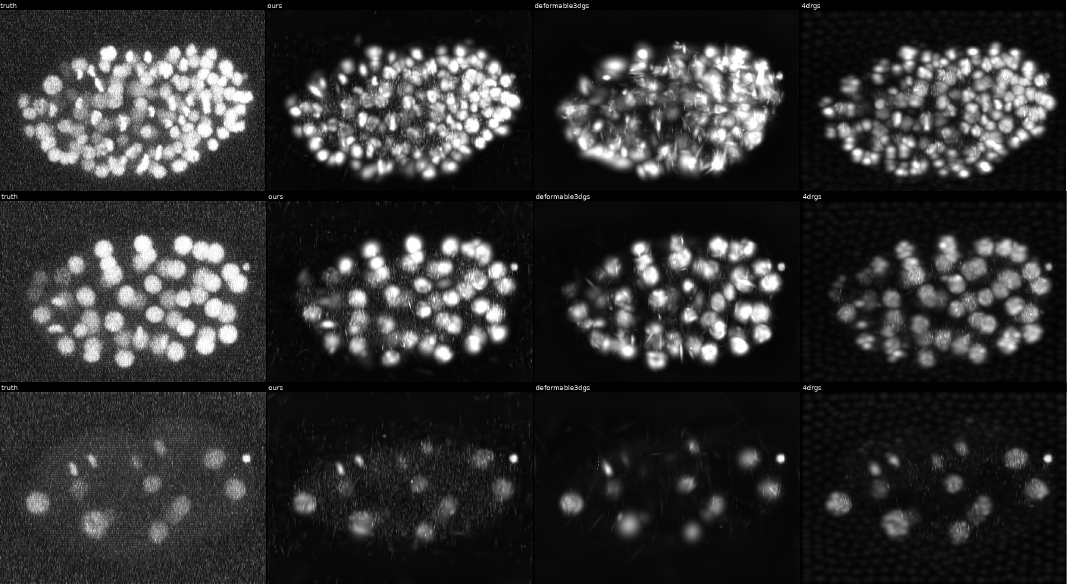}
\caption{Qualitative comparison across the acquisition gap at $k=8$ (CE02).
Fidelity next to an acquired frame is not what separates the methods; what
separates them is how much structure survives to the middle of the gap.}
\label{fig:comparison}
\end{figure*}

\subsection{Two further protocol choices}

\textbf{Flux is never reported alone.} The learnable-$\sigf$ experiment
(Sec.~\ref{sec:psf}) is a worked example of it being gamed: $+0.176$ flux for
$-1.18$~dB held-out axial PSNR. Every flux number here is paired with a
held-out metric.

% --- comparison.md main table, restricted to the five comparison arms;
% --- decay percentages recomputed from the d=1 and d=4 columns --------------
\begin{table*}[!tb]
\caption{Held-out nucleus flux at $k=8$ (CE02), by distance $d$ to the nearest
acquired frame. Clean role, quiescent phase, frozen cross-arm panel, median of
three seeds. \textbf{The primary statistic is the relative decay from $d{=}1$
to $d{=}4$}.}
\label{tab:main}
\centering
\footnotesize
\begin{tabular}{lcccccccc}
\toprule
Method & $d{=}$all\hi & $d{=}1$\hi & $d{=}2$\hi & $d{=}3$\hi & $d{=}4$\hi &
\textbf{Decay $d{=}1\!\rightarrow\!4$}\hi & Division window\hi & Primed\hi \\
\midrule
\textbf{Ours} & \textbf{0.481} & \textbf{0.694} & \textbf{0.476} & \textbf{0.326} & \textbf{0.370} & $\mathbf{-46.7\%}$ & \textbf{0.417} & \textbf{0.664} \\
Deformable 3DGS \cite{yang2024deformable} & 0.194 & 0.439 & 0.167 & 0.156 & 0.101 & $-77.0\%$ & 0.263 & 0.468 \\
4DRGS \cite{liu20244drgs}                 & 0.233 & 0.576 & 0.290 & 0.152 & 0.080 & $-86.1\%$ & 0.303 & 0.432 \\
\bottomrule
\end{tabular}
\end{table*}

\textbf{Centroid offset is in absolute pixels, not a ratio.} The observation's
own offset against the same mask is $0.71$--$0.84$~px; using that as a
denominator turns a sub-pixel measurement error into the divisor, so a render
misplaced by $4$~px reads as ``$5\times$''. The absolute statement is the
physically meaningful one.

% --- comparison.md tables 2 and 4 -------------------------------------------
\begin{table}[!tb]
\caption{Gating and morphology at $k=8$. Sharpness is measured on held-out
frames; centroid offset is in absolute pixels.}
\label{tab:gating}
\centering
\footnotesize
\setlength{\tabcolsep}{2pt}
\begin{tabular}{lcccc}
\toprule
Method & Train flux\hi & Sharpness\hi & Retention\hi & Centroid\lo \\
\midrule
\textbf{Ours}   & 0.878 & \textbf{0.225} & \textbf{0.547} & 14.13 \\
Deformable 3DGS & \textbf{0.908} & 0.101 & 0.214 & 22.81 \\
4DRGS           & 0.906 & 0.088 & 0.258 & \textbf{12.08} \\
\bottomrule
\end{tabular}
\end{table}

% ---------------------------------------------------------------------------
\begin{table}[!tb]
\caption{Axial generalization on two sequences, as masked $z$-PSNR (dB). All
share the PSF-aware forward model, so these columns compare temporal representations.
Two columns come from \emph{same} volumes under \emph{same} foreground mask and differ only in which $z$-planes they pool:
\emph{sup.}\ covers the planes the model was fitted to, \emph{held} the planes
removed by the axial holdout and never seen in training.}
\label{tab:cost}
\centering
\footnotesize
\setlength{\tabcolsep}{4pt}
\begin{tabular}{lcccc}
\toprule
\textbf{Masked} $z$-PSNR (dB)\hi & \multicolumn{2}{c}{CE02, $k{=}8$}
  & \multicolumn{2}{c}{CHO, $k{=}4$} \\
\cmidrule(lr){2-3}\cmidrule(lr){4-5}
Method & sup. & held & sup. & held \\
\midrule
\textbf{Ours}   & \textbf{12.00} & \textbf{12.04} & 15.94 & \textbf{12.23} \\
Deformable 3DGS & 11.12 & 11.16 & 14.58 & \phantom{0}9.01 \\
4DRGS           & 11.49 & 11.40 & \textbf{16.81} & 11.70 \\
\bottomrule
\end{tabular}
\end{table}

\section{Experiments}
\label{sec:exp}

\subsection{Setup}

Flux and ablation results use CE02 at $k=8$ with a shared frozen evaluation
panel, three seeds, and the same held-out frames and masks for every arm;
Table~\ref{tab:cost} adds CHO at $k=4$ under the same protocol. We compare
against Deformable 3DGS \cite{yang2024deformable}, a deformation-field method,
and 4DRGS \cite{liu20244drgs}, a native-4D method in the same emission-only
regime as ours (Sec.~\ref{sec:prelim}). Every method is refitted from scratch
at both strides under its own published budget policy, with Adam
\cite{kingma2015adam}; none inherits a model from a denser setting.

\textbf{All arms render through the same PSF-aware forward model}
(Sec.~\ref{sec:whypsf}) with the same measured $\sigf$, not each method's
published optics, so no baseline is penalized for the axial blur its original
formulation leaves free and the comparison isolates how each represents
\emph{time}. The PSF's own value is established in Sec.~\ref{sec:psf}.

% ---------------------------------------------------------------------------
\begin{figure}[!tb]
\centering
\includegraphics[width=\columnwidth]{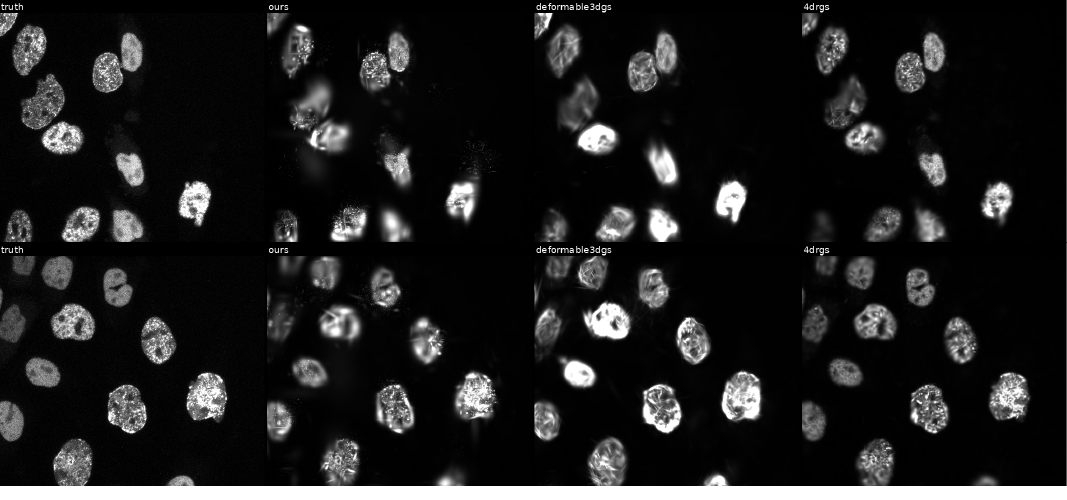}
\caption{Second cell line (CHO, $k{=}4$). Held-out $z$-planes rendered by each
method against the acquired volume, at the top and bottom of the stack where a
plane has neighbours on one side only. The sequence transfers without retuning:
no hyper-parameter is changed from the CE02 setting except the gate
initialization, which follows the stride.}
\label{fig:cho}
\end{figure}

% ---------------------------------------------------------------------------
\begin{figure*}[!tb]
\centering
\includegraphics[width=\textwidth]{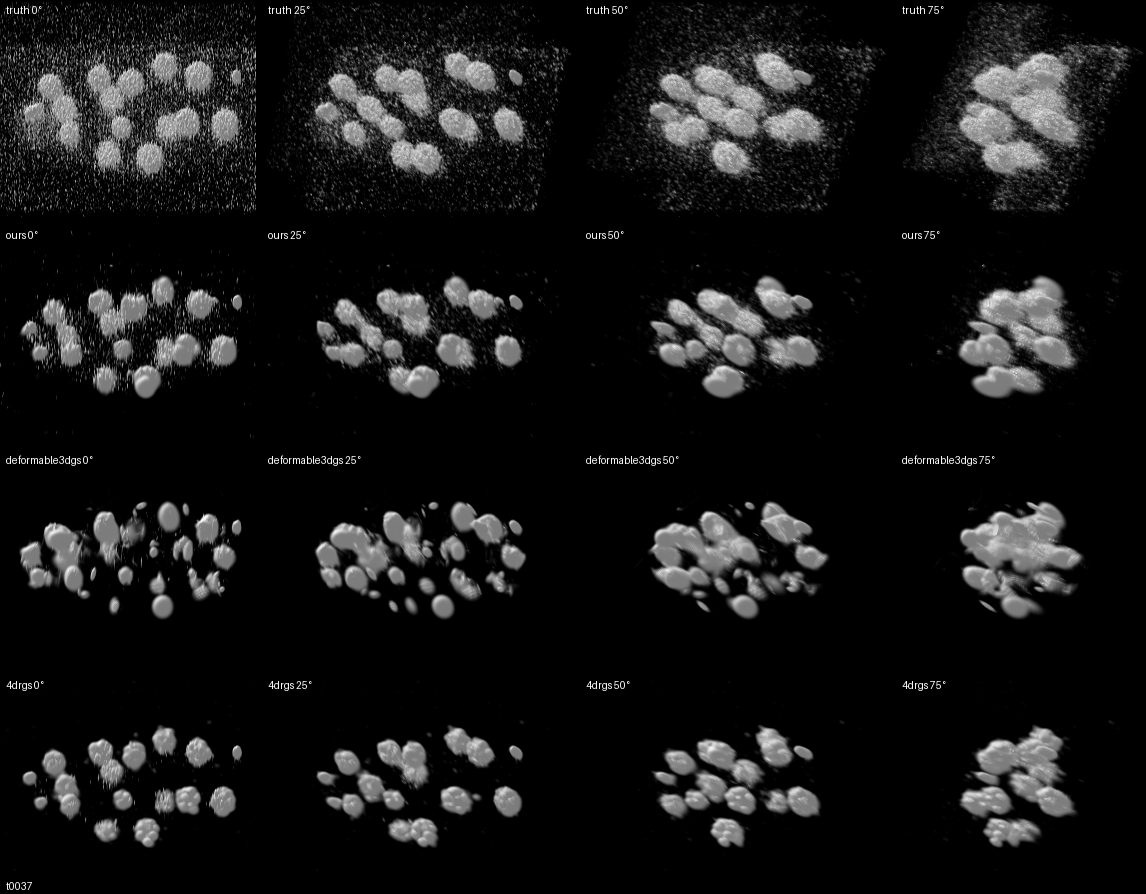}
\caption{Volume renderings of one held-out timestep in the middle of the gap
($d{=}1$, $k{=}8$, CE02): the acquired volume and the three reconstructions.
The slice views of Fig.~\ref{fig:comparison} show what survives in plane;
rendered as volumes, what separates the methods is how many nuclei stay
resolvable as separate bodies instead of merging into their neighbours or
disappearing.}
\label{fig:volume}
\end{figure*}

\subsection{Main result: graceful degradation across the gap}

Table~\ref{tab:main} is the central result; Fig.~\ref{fig:comparison} shows it
in plane and Fig.~\ref{fig:volume} as volumes. Ours has the flattest decay across the gap, $-46.7\%$
against $-77.0\%$ for the next best, and the highest fidelity at every distance
from an acquired frame, including the midpoint of the gap, the hardest timestep
the robot ever has to act on. Any single cell is within the spread one method shows across seeds, so the
slope, pooling all four intervals, is what we rest the claim on.

\textbf{The shape of the curve is the argument, not the height of any one point
on it.} Beside an acquired frame the methods are close, our lead at $d{=}1$
being a factor of $1.2$ over the best baseline; by the middle of the gap it is
a factor of $3.7$. A method that is adequate at $d{=}1$ and steep thereafter
has learned to smooth between observations rather than to model how the
specimen moves, and buys a robot nothing. Holding fidelity far from an
acquired frame is what the acquisition schedule rewards, because a flat profile
turns a longer interval between acquisitions from lost information into saved
photon budget.

% --- ablation.json (prefix ladder_k8) ---------------------------------------
\begin{table}[!tb]
\caption{Completion-signal ablation, $k=8$, CE02, as mean $\pm$ s.d.\ over three seeds. \emph{off}
supervises only the acquired frames; $\Tint$ adds a low-frequency
pseudo-target at the unobserved timesteps.}
\label{tab:ladder}
\centering
\footnotesize
\setlength{\tabcolsep}{2.2pt}
\begin{tabular}{lccccc}
\toprule
Variant & Flux\hi & Centroid\lo & Div.\ win.\hi & Train\hi & Axial\tgt{0} \\
    & (clean) & (px) & flux & flux & gap \\
\midrule
off   & $0.353${\scriptsize$\pm$.068} & $\mathbf{10.49}${\scriptsize$\pm$.87} & $0.439${\scriptsize$\pm$.111} & $0.892${\scriptsize$\pm$.033} & $0.171${\scriptsize$\pm$.025} \\
$\Tint$ & $\mathbf{0.480}${\scriptsize$\pm$.042} & $11.10${\scriptsize$\pm$.84} & $\mathbf{0.449}${\scriptsize$\pm$\textbf{.012}} & $0.890${\scriptsize$\pm$.023} & $\mathbf{0.097}${\scriptsize$\pm$.021} \\
\bottomrule
\end{tabular}
\end{table}

\subsection{Gating, morphology and axial generalization}

A decay slope can be flattened two ways: by extrapolating better, or by fitting
the acquired frames worse. \emph{Retention}, held-out fidelity over training
fidelity, tells them apart (Table~\ref{tab:gating}). It is mostly the first: ours fits the acquired frames marginally less well than
either baseline ($0.878$ against $0.908$ and $0.906$) yet retains \textbf{more
than twice as much of that fit} on frames it never saw ($0.547$ against $0.214$
and $0.258$), so the flat slope is not underfitting. Held-out
sharpness orders the same way; centroid offset is the one place we trail.

Axially (Table~\ref{tab:cost}), ours and Deformable 3DGS score the same on
planes they were fitted to and planes they were not, while 4DRGS scores worse
on the withheld ones, propped up by the planes it saw, which is the failure
Eq.~\eqref{eq:sum} anticipates when axial extent is left to the primitives. CHO is the harder test and no method passes it cleanly (Fig.~\ref{fig:cho}),
because its holdout withholds a larger fraction of planes and the outermost
ones; held-out scores are read down a column, never across sequences. However,
\textbf{Ours reconstructs planes it never saw better than any baseline on
either sequence} ($12.04$ on CE02, $12.23$ on CHO). We do not read the
agreement between a method's two columns as evidence for the PSF: every arm
renders through the same one, and Deformable 3DGS still matches our CE02 temporal representation gap. Whole-volume PSNR and SSIM are not tabulated: they place our arm and 4DRGS
within $5\%$ of each other and disagree about which is ahead on \emph{both}
sequences, where held-out flux separates the two more than twofold. Their one
large spread, $24\%$ in CHO PSNR, only restates that Deformable 3DGS is worst.
Sec.~\ref{sec:psnrfails} predicts both behaviours of an unmasked metric on
$92\%$ background.

\subsection{Ablation: what the completion signal is worth}
\label{sec:attribution}

To isolate the completion signal we compare \emph{off}, in which the unobserved
timesteps receive no supervision, against $\Tint$, in which they receive the
completion network's output, at $k=8$ on the frozen clean/quiescent panel
($n=58$) with everything else held fixed (Table~\ref{tab:ladder}). Division-window flux
restricts the statistic to nuclei caught mid-division, the events a lineage
graph exists to record.

\textbf{Injection raises held-out fidelity} with no seed-wise overlap between
the groups, and does so precisely where it supervises: the two variants are
indistinguishable on the acquired frames they both saw, so the gain is
localized to the unobserved timesteps rather than to a generally better fit. \textbf{It also closes the axial gap}, by more than two standard deviations,
which is not obvious since the pseudo-targets carry no axial information the
acquired frames did not already have: constraining the model in time evidently
constrains it in depth. \textbf{And it makes
the outcome reproducible}, which for a platform retraining unattended matters more than any mean: on division-window flux the \emph{means} are nearly
identical while the spreads differ ninefold ($\pm 0.012$ against $\pm 0.111$), so without injection whether a run resolves dividing
nuclei at all is decided by the seed.

% ---------------------------------------------------------------------------
\FloatBarrier
\section{Conclusion}

A robotic microscope is limited by its acquisition schedule, not by its
analysis software: the volumes a photon budget will not pay for are the ones
its downstream tools assume it has. We have treated those missing frames as a
completion problem, fitting a continuous-time 4D state that renders any
timestep on demand for the segmentation, tracking and lineage software a lab
already runs. Three changes make 4D splatting work at this sampling rate: a temporal gate
measured in frames rather than as a fraction of the recording; an axial PSF
pinned to a physical constant, which removes a degeneracy the optimizer would
otherwise exploit and argues for spending a tight budget on time rather than
depth; and gate-aware densification under a budget. What they buy is a reconstruction that degrades gently across the gap
rather than only beside the frames that were acquired, and that recovers focal
planes it was never shown. Graceful degradation, not peak fidelity, turns a longer acquisition interval
into saved photon budget. The natural next step closes the loop in the other direction: using the
reconstruction's own per-timestep uncertainty to tell the robot \emph{when} the
next volume is worth acquiring, as uncertainty-driven view selection already
does for \emph{where} to look next \cite{pan2022activenerf}.

% ---------------------------------------------------------------------------

\end{document}